\pdfoutput=1

\documentclass[11pt]{article}

\usepackage{acl}
\usepackage{annotates}
\usepackage{times}
\usepackage{latexsym}

\usepackage[T1]{fontenc}

\usepackage[utf8]{inputenc}

\usepackage[english]{babel}
\usepackage[english=british]{csquotes}

\usepackage{microtype}

\usepackage{graphicx}
\usepackage{booktabs}
\usepackage{float}

\title{Word-order typology in Multilingual BERT: \\ A~case study in subordinate-clause detection}

\author{Dmitry Nikolaev \qquad Sebastian Pad{\'o} \\
  IMS, University of Stuttgart \\
  \texttt{dnikolaev@fastmail.com}, \texttt{pado@ims.uni-stuttgart.de}
  }

\begin{document}
\maketitle

\begin{abstract}
  The capabilities and limitations of BERT and similar models are
  still unclear when it comes to learning syntactic abstractions, in
  particular across languages. In this paper, we use the task of
  \textit{subordinate-clause detection} within and across languages to
  probe these properties. We show that this task is deceptively
  simple, with easy gains offset by a long tail of harder cases, and
  that BERT's zero-shot performance is dominated by word-order effects,
  mirroring the SVO/VSO/SOV typology.
\end{abstract}

\section{Introduction}

Analysing the ability of pre-trained neural language models, such as
BERT \citep{devlin-etal-2019-bert}, to abstract grammatical patterns
from raw texts has become a prominent research question
\citep{jawahar-etal-2019-bert,rogers-etal-2020-primer}. 
Results remain mixed. While BERT-based models have been shown to learn
syntactic representations that are similarly structured across
languages \citep{chi-etal-2020-finding}, some grammatical patterns,
such as discontinuous constituents, remain challenging for them even
when training data is plentiful
\citep{kogkalidis2022discontinuous}. In practical terms, zero-shot
performance of BERT-based models is lower for typologically distant
languages \citep{pires-etal-2019-multilingual}, and they can profit
from direct exposure to typological features during fine-tuning
\citep{bjerva-augenstein-2021-typological}.

In this study, we add another datapoint to the conversation by
analysing the ability of BERT-based models to capture the
\textit{distinction between main and subordinate clauses} across
languages. This task is promising for two reasons. First, it
highlights variability in the way main and subordinate clauses are
structured across languages, thus acting as an informative probe into
the relationship between BERT and typological categories. Second, the
task is arguably relevant for downstream performance on 
natural-language understanding, where (some notion of) syntactic scope and
compositionality should support tasks such as analysing
commitment
\citep{jiang-de-marneffe-2019-evaluating,zhang-de-marneffe-2021-identifying}
or factuality \cite{lotan-etal-2013-truthteller}, text simplification
\citep{sikka2020survey}, or paraphrase detection
\citep{timmer2021can}. In order to operationalise it in a
cross-lingual fashion, we use the Universal Dependencies framework
\citep[UD;][]{nivre-etal-2020-universal} with its large 
multilingual collection of corpora.

Our analysis proceeds in two stages. First, we survey the performance
of BERT models fine-tuned and tested on the same language across 20
typologically diverse languages (\S~\ref{sec:single-language}). For
the majority of languages, distinguishing main and subordinate clauses
is easily solved with base-size models and relatively small training
sets. However, some languages demonstrate a non-negligible number of
errors, which we analyse.

Then we study the performance of Multilingual BERT (mBERT) in a
zero-shot setting (\S~\ref{sec:zero-shot}), where we fine-tune the
model on labeled data in 10 different languages and then test its
performance on 31 datasets representing 27 different languages. We
find that the performance of mBERT is dominated by word-order effects
well known from the typological literature \cite{comrie81:_languag}:
the Arabic model shows best-in-class performance on Irish, and the
Japanese model has best-in-class performance on Korean, while both
have poor performance overall. European languages with large training
sets provide good inductive bias for typologically diverse languages
but fail on SOV languages.

\begin{table*}[th]
  \centering
  \setlength\tabcolsep{2pt}
\resizebox{\textwidth}{!}{
\begin{tabular}{lllllllllll}
\toprule
\textbf{Language} & Mandarin & Vietnamese & Korean & Arabic & Hindi & German & Armenian & Turkish & Welsh & Indonesian \\
\textbf{Accuracy} & 88.7 & 90 & 90.4 & 91.2 & 93.6 & 94.1 & 94.3 & 95.1 & 95.6 & 96 \\ \midrule
\textbf{Language} & Basque & Spanish & Irish & English & Hebrew & Afrikaans & French & Japanese & Czech & Russian \\
\textbf{Accuracy} & 96.9 & 97.1 & 97.4 & 97.9 & 98.2 & 98.8 & 99 & 99.1 & 99.6 & 99.7 \\
\bottomrule
\end{tabular}}
\caption{Performance of single-language models.}\label{tab:single-lang-small}
\end{table*}

\section{Experimental Setup}\label{sec:setup}

\paragraph{Data}

To make our analysis maximally comparable across languages, we start
from the Parallel Universal Dependencies (PUD) collection
\citep{zeman-etal-2017-conll}, which contains translations for a set
of 1000 English sentences. PUD only contains test corpora. As these
are too small to be further split into train/test subsets, we use
other corpora to fine-tune the models. We also add corpora for
languages not covered by PUD for better typological coverage.  See
Appendix \S~\ref{ssec:corpora} for the full list.

\paragraph{Model}

The experimental setup is identical in the single-language and
zero-shot settings. A~pre-trained mBERT model (a
variant of \texttt{bert-base}) and several pre-trained single-language
BERT models, all provided by
HuggingFace \citep{wolf-etal-2020-transformers}, are fine-tuned on
the binary classification of predicates into main vs.\ subordinate
clauses. We operationalize main clauses as those headed by predicates
with the UD label \texttt{root} and subordinate clauses through the UD labels
\texttt{acl}, \texttt{ccomp}, \texttt{advcl}, \texttt{csubj}, and
\texttt{xcomp}. The last hidden state of the embedding model for the
first subword of each predicate is fed to a two-layer MLP with a tanh
activation after the first layer, and the model is fine-tuned using
cross-entropy loss. For the single-language setup, the model is
fine-tuned for five epochs, and we report the best result on the
validation set. Most models begin overfitting after the second epoch,
so in the zero-shot setting all models are fine-tuned for two epochs.

\section{Single-Language Models}\label{sec:single-language}

The main results obtained by the models fine-tuned and tested on the
same language are shown in Table~\ref{tab:single-lang-small}. Results
are above 90\% for almost all languages, while a majority baseline
(always assign subordinate clause) attains an accuracy of 50--70\%
depending on the language. (Table~\ref{tab:single-lang-full} in the
Appendix provides more details about the models and corpora, including
exact baseline results.) At first glance, neither the size of the
training set nor the size of the model seem to be a major factor:
mBERT demonstrates better performance when fine-tuned on the small
Afrikaans and Hebrew datasets than when trained on a bigger Chinese
dataset. When fine-tuned on the English data, it attains the same
performance as an English-only \texttt{bert-large}.\footnote{The mBERT
result is reported in Table~\ref{tab:zero-shot}.}

A~more fundamental distinction seems to exist between major European
languages, the results on which are generally at $>$ 97\% accuracy
(except for German), and Mandarin Chinese, Vietnamese, and Korean where
results are around 90\%. Our analysis indicates that these differences
are partly due to discrepancies in UD annotations across corpora but
also due to genuine syntactic differences. An example of an
annotation-related confound is the treatment of quotations. The PUD
corpora that we use preferentially as test sets treat quotations as 
sentential complements of communication verbs. Some of the corpora we use for 
fine-tuning, however, analyse the cases where quotation precedes the verb 
of speech as parataxis. The head predicate of the quotation therefore receives
the label \texttt{root} and becomes the main predicate of the whole
sentence, leading to spurious mistakes in the analysis of PUD corpora,
where they are annotated as \texttt{ccomp}'s. This discrepancy
accounts for the lion's share of classification mistakes in German and
some mistakes in Mandarin.

In contrast, an example of genuine ambiguity is provided by the
Mandarin \textit{gēnjù} construction. This construction means
\enquote{according to} and can incorporate both nominal and verbal
constituents. Thus, \textit{gēnjù shàng biǎogé zhōng qī gè yuánsù de
guānxì} from the Mandarin GSD corpus, which we used for fine-tuning,
means \enquote{based on the relationship of the seven elements in the above table}, 
and the annotation treats this construction as an oblique prepositional phrase. 
Cf.\ the following example from the Mandarin PUD corpus: \textit{gēnjù kěxíng xìng
yánjiū gūjì} \enquote{according to the feasibility study / the
feasibility study estimates that / as the feasbility study
estimates}. The analysis of this sentence in PUD makes \textit{gūjì}
\enquote{estimate} the main predicate of the sentence, while an
alternative analysis would make it the head of an adverbial clause,
and yet another analysis would label it as a nominal element. The
ability of Mandarin words to act as different parts of speech in
different contexts (especially in case of verbs, which can act as
clause heads, auxiliaries, complementisers, and compound elements)
makes this kind of disambiguation difficult even for human annotators, which
in turn makes it hard to formulate the exact rule that language models
are supposed to extract from the data. A~similar situation holds for
Vietnamese.\footnote{\enquote{Syntactic category classification for
Vietnamese is still in debate. That lack of consensus is due to the
unclear limit between the grammatical roles of many words as well as
the frequent phenomenon of syntactic category mutation}
\citep{nguyen2004vietnamese}.}

A different type of systematic ambiguity is presented by Korean, which
also demonstrates poorer performance. Korean has about sixty markers
connecting two clauses, and many of those allow for both coordinative
and conjunctive readings, which makes either the first or
the second clause the main one, respectively
\citep[220--227]{cho2020korean}. Examples of this type are responsible
for a large share of mistakes in Korean.

Overall, these results indicate that subordinate-clause detection is a
long-tail task: major easily learnable patterns account for more than
90\% of test cases for all languages, but in some languages there is
an assortment of harder cases that prevent language models from
efficiently generalising.

\section{Zero-Shot Setting}\label{sec:zero-shot}

\begin{table*}[h]
\centering
\resizebox{\textwidth}{!}{
\begin{tabular} {lrrrrrrrrrrrrrrrr}
\toprule
 & \textbf{\begin{tabular}[c]{@{}l@{}}ar\\ padt\end{tabular}} & \textbf{\begin{tabular}[c]{@{}l@{}}ga\\ idt\end{tabular}} & \textbf{\begin{tabular}[c]{@{}l@{}}af\\ booms\end{tabular}} & \textbf{\begin{tabular}[c]{@{}l@{}}de\\ pud\end{tabular}} & \textbf{\begin{tabular}[c]{@{}l@{}}cs\\ pud\end{tabular}} & \textbf{\begin{tabular}[c]{@{}l@{}}cy\\ ccg\end{tabular}} & \textbf{\begin{tabular}[c]{@{}l@{}}en\\ ewt\end{tabular}} & \textbf{\begin{tabular}[c]{@{}l@{}}en\\ pud\end{tabular}} & \textbf{\begin{tabular}[c]{@{}l@{}}es\\ pud\end{tabular}} & \textbf{\begin{tabular}[c]{@{}l@{}}fi\\ pud\end{tabular}} & \textbf{\begin{tabular}[c]{@{}l@{}}fr\\ pud\end{tabular}} & \textbf{\begin{tabular}[c]{@{}l@{}}he\\ hdt\end{tabular}} & \textbf{\begin{tabular}[c]{@{}l@{}}hy\\ arm\end{tabular}} & \textbf{\begin{tabular}[c]{@{}l@{}}id\\ pud\end{tabular}} & \textbf{\begin{tabular}[c]{@{}l@{}}is\\ pud\end{tabular}} & \textbf{\begin{tabular}[c]{@{}l@{}}it\\ pud\end{tabular}} \\
\midrule
English & \textbf{96} & 95 & 93 & 94 & 94 & 86 & \textbf{98} & \textbf{98} & \textbf{96} & 96 & 96 & 93 & 93 & \textbf{95} & \textbf{96} & \textbf{98} \\
Russian & 95 & 94 & 86 & 93 & 95 & 90 & 94 & 96 & 95 & \textbf{97} & \textbf{99} & 94 & \textbf{94} & 93 & \textbf{96} & 96 \\
Czech & 94 & 94 & 83 & \textbf{95} & \textbf{100} & \textbf{92} & 92 & 93 & 94 & 94 & 98 & \textbf{95} & 88 & 89 & 92 & 94 \\
French & 94 & 91 & 84 & 92 & 95 & 90 & 92 & 97 & 95 & 95 & \textbf{99} & 94 & 87 & 93 & \textbf{96} & 96 \\
German & 94 & 87 & \textbf{95} & 94 & 88 & 82 & 90 & 95 & 94 & 92 & 93 & 91 & 90 & 89 & 95 & 96 \\
Arabic & 90 & \textbf{96} & 76 & 85 & 84 & 90 & 85 & 87 & 86 & 85 & 84 & 87 & 84 & 85 & 89 & 85 \\
Mandarin & 86 & 84 & 85 & 87 & 87 & 85 & 85 & 86 & 86 & 89 & 87 & 87 & 81 & 83 & 86 & 87 \\
Turkish & 67 & 61 & 61 & 65 & 71 & \underline{62} & 64 & 69 & 75 & 77 & 71 & 73 & 73 & 68 & 68 & 74 \\
Korean & \underline{51} & \underline{53} & 63 & \underline{53} & 61 & \underline{52} & \underline{51} & \underline{54} & \underline{59} & 65 & \underline{59} & 59 & \underline{57} & \underline{55} & \underline{54} & 61 \\
Japanese & \underline{55} & \underline{56} & \underline{41} & \underline{39} & \underline{52} & \underline{63} & \underline{51} & \underline{52} & \underline{52} & \underline{54} & \underline{55} & \underline{54} & \underline{58} & \underline{53} & \underline{51} & \underline{54} \\
\midrule
 & \textbf{\begin{tabular}[c]{@{}l@{}}pl\\ pud\end{tabular}} & \textbf{\begin{tabular}[c]{@{}l@{}}pt\\ pud\end{tabular}} & \textbf{\begin{tabular}[c]{@{}l@{}}ru\\ pud\end{tabular}} & \textbf{\begin{tabular}[c]{@{}l@{}}ru\\ syntag\end{tabular}} & \textbf{\begin{tabular}[c]{@{}l@{}}sv\\ pud\end{tabular}} & \textbf{\begin{tabular}[c]{@{}l@{}}eu\\ bdt\end{tabular}} & \textbf{\begin{tabular}[c]{@{}l@{}}hi\\ pud\end{tabular}} & \textbf{\begin{tabular}[c]{@{}l@{}}tr\\ pud\end{tabular}} & \textbf{\begin{tabular}[c]{@{}l@{}}ja\\ gsd\end{tabular}} & \textbf{\begin{tabular}[c]{@{}l@{}}ja\\ pud\end{tabular}} & \textbf{\begin{tabular}[c]{@{}l@{}}ko\\ pud\end{tabular}} & \textbf{\begin{tabular}[c]{@{}l@{}}vi\\ vtb\end{tabular}} & \textbf{\begin{tabular}[c]{@{}l@{}}th\\ pud\end{tabular}} & \textbf{\begin{tabular}[c]{@{}l@{}}zh\\ gsd\end{tabular}} & \textbf{\begin{tabular}[c]{@{}l@{}}zh\\ pud\end{tabular}} & \textbf{mean} \\
\midrule
English & 95 & \textbf{97} & 93 & 93 & 96 & \textbf{88} & 87 & 83 & \underline{66} & \underline{70} & 67 & \textbf{82} & \textbf{84} & \underline{67} & 71 & 88.9 \\
Russian & \textbf{98} & 95 & \textbf{100} & \textbf{99} & 95 & \textbf{88} & 90 & 90 & \underline{68} & 72 & 70 & 79 & 80 & \underline{64} & 69 & \textbf{89.2} \\
Czech & 97 & 93 & 94 & 96 & 92 & 87 & 88 & 88 & \underline{64} & \underline{66} & 68 & 78 & 79 & \underline{65} & 71 & 87.5 \\
French & \textbf{98} & 96 & 96 & 97 & 94 & 85 & \textbf{89} & 86 & \underline{54} & \underline{61} & 66 & 77 & 76 & \underline{63} & 69 & 87.0 \\
German & 89 & 94 & 93 & 88 & \textbf{97} & 81 & 86 & 78 & \underline{59} & \underline{62} & \underline{57} & 78 & 78 & \underline{67} & 68 & 85.2 \\
Arabic & 85 & 86 & 88 & 85 & 89 & 71 & 70 & 65 & \underline{63} & \underline{66} & \underline{59} & 74 & 79 & \underline{66} & \underline{65} & 80.3 \\
Mandarin & 85 & 85 & 86 & 85 & 86 & 82 & 87 & 89 & 80 & 78 & 77 & 74 & 80 & \textbf{91} & \textbf{86} & 84.6 \\
Turkish & 76 & 73 & 71 & 71 & 69 & 79 & 83 & \textbf{94} & 82 & 83 & 88 & \underline{63} & \underline{68} & 72 & 71 & 72.3 \\
Korean & 66 & \underline{58} & \underline{59} & \underline{59} & \underline{53} & 74 & 76 & \textbf{94} & 87 & 88 & 88 & \underline{52} & \underline{61} & \underline{67} & \underline{66} & 63.1 \\
Japanese & \underline{54} & \underline{52} & \underline{55} & \underline{55} & \underline{50} & \underline{57} & 63 & 88 & \textbf{99} & \textbf{98} & \textbf{95} & 54 & \underline{70} & 72 & \underline{66} & 60.3 \\
\bottomrule
\end{tabular}}
\caption{Performance of zero-shot models. Rows: source languages;
  columns: target languages and corpora. Underlined values fail to
  beat the majority-class baseline (always predict subordinate
  clause). See \S~\ref{ssec:abbreviations} for language abbreviations
  and \S~\ref{ssec:corpora} for details about
  corpora.}\label{tab:zero-shot}
\end{table*}

\subsection{Quantitative Results}\label{ssec:zs-quantitative}

We now turn to the analysis of the performance of the models in the
zero-shot setting. The model described in \S~\ref{sec:setup} is
fine-tuned for two epochs on five European languages (English,
Russian, Czech, French, and German) and five Eurasian languages
(Standard Arabic, Mandarin Chinese, Turkish, Korean, and Japanese)
with larger training corpora (the ones shown in
Table~\ref{tab:single-lang-full}). Each of the fine-tuned models is
then applied in a zero-shot way to a range of test corpora from the UD
collection.\footnote{Where available, we experiment with two test sets for
  the same language to assess domain-induced variance.  As
  Table~\ref{tab:zero-shot} shows, the difference in scores between
  different testing corpora for the same language can reach
  5--6\%, but it does not change the overall pattern.}

Based on the results in Table~\ref{tab:zero-shot}, several
observations can be made. First, there is a set of European languages
with large training corpora that can act as \enquote{general
approximators}: they demonstrate high performance across the
board. The best overall performance is attained by Russian, which has
the second-largest training corpus (nearly 33k sentences).  German,
with the largest training corpus (nearly 56k sentences) performs worse
than both Russian and English (the second best, with only circa 6k
training sentences). While this good result for English may be
attributed to more informative pre-training (English Wikipedia is much
larger than the German one), such a bias would also have favoured
German compared to Russian. An alternative explanation is provided by
the more idiosyncratic German word-order patterns (V2 in main clauses
vs.\ V-last in subordinate clauses), which help it achieve
best-in-class performance on the similar Afrikaans.  Notably, Russian
beats English even though PUD corpora were translated from English and
therefore should contain some traces of its morphosyntactic patterns
\citep{rabinovich-etal-2017-found,nikolaev2020morphosyntactic}.

At the other end of the spectrum, we find mediocre general
approximators (Arabic, Turkish) and outright bad ones (Japanese and
Korean). At first glance, their performance could be an artefact of
lower-quality annotations or suboptimal tokenisation
\citep{mielke2021between}. This, however, does not explain a
remarkable set of results that is clearly due to word-order patterns.
While the fine-tuned model for Arabic, a VSO language, performs worse
on its own test corpus than models fine-tuned on European languages,
it provides best-in-class performance on Irish, another VSO language
(96\% accuracy). The English-based model is not far behind (95\%), but
given the overall large gap in performance between them across the
board, it seems that congruent word-order patterns provide a
strong inductive bias for subordinate-clause identification.

Unfortunately, VSO languages are rare,\footnote{Out of 1376 languages
  in WALS \citep{wals}, 95 are VSO, 564 are SOV and 488 are SVO.} and
  it is impossible to check if this pattern generalises to other
  language pairs. However, our test-corpus suite includes data on
  strict SOV languages (Japanese, Korean) and languages where SOV is
  the dominant (Hindi, Turkish) or a common (Mandarin, Basque)
  pattern. These provide us with a large number of language pairs with
  different degrees of word-order congruence and fairly clear patterns
  of model performance. First, universal approximators, despite good
  performance on VSO languages, struggle on strict SOV languages,
  especially Japanese, while SOV languages demonstrate consistently
  good performance among themselves. E.g., Korean demonstrates
  best-in-class performance on Turkish, tied with Turkish itself,
  while Japanese has best-in-class performance on Korean. Turkish also
  demonstrates decent performance on Hindi, with which it shares a
  relatively flexible SOV order.

Another language with 
strong SOV tendencies is Mandarin Chinese, which has been argued to be
in transition from SVO to SOV order \citep{sun1985sov}. Mandarin,
which we already found difficult to model in
\S~\ref{sec:single-language}, is very hard to generalise to, with no
source languages attaining accuracy above 71--72\%. Tellingly,
Turkish is the only other language with decent results on both
Mandarin test sets. Mandarin is also the only language to always beat
the majority-class baseline.

\subsection{Case Study: English--Mandarin}\label{ssec:case-study}

In order to get a better understanding of the difficulties that models
face in the zero-shot setting we analysed the mistakes that the English-based
fine-tuned model made when making predictions on Mandarin data.

Setting aside errors stemming from annotation
discrepancies,\footnote{E.g., as discussed in \S 3, the model expects
direct quotes to have the form \texttt{ccomp} (quote) $+$
\texttt{root} (verb of speech) and not \texttt{root} (quote) $+$
\texttt{parataxis} (verb of speech).} the major source of model
mistakes seems to be the fact that Mandarin complex sentences are
predominantly right-headed: 99\% of \texttt{advcl}, 100\% of
\texttt{acl}, and 96\% of \texttt{dep}\footnote{\texttt{dep} labels
different kinds of hard-to-analyse relations and is frequent in
Mandarin PUD (397 occurrences).}  have their parent node to the
right. In contrast, 75\% of English \texttt{advcl} and 98\% of English
\texttt{acl} are left-headed in PUD. This makes an English-based
zero-shot model prejudiced against finding \texttt{root} nodes in the
final clause of the sentence, and it incorrectly analyses a wide range
of right-headed Mandarin complex clauses. Statistically, there are 142
sentence-initial subordinate clauses mistakenly analysed as main
clauses and only 6 reverse errors. By contrast, there are 278
sentence-final main clauses mistakenly analysed as subordinate ones
and 82 reverse errors.

Sometimes this divergence further interacts with ways in which English
and Mandarin alternate between clause coordination and
subordination. Thus, Mandarin tends to describe sequences of events as
a pair of an adverbial clause and a main clause (\textit{after having
taken a shower, he dried himself}) instead of as two coordinated
clauses (\textit{he took a shower and dried himself}). English UD
treats the first conjoined clause as the matrix one, while it is often
\texttt{advcl} in Mandarin, and the absence of overt unambiguous
complementisers makes it hard for the model to see beyond mere
frequencies.

A~similar situation obtains with some English postposed descriptive subordinate 
clauses, such as \textit{it's X--that Y} 
constructions\footnote{\textit{It's fantastic \underline{that they got the
Paris Agreement}, but...}} and non-restrictive relative
clauses.\footnote{\textit{However, they could not find this same
pattern in tissues such as the bladder, \underline{which are not
directly exposed}.}} In these cases, Mandarin uses a coordinative
construction, in which the head, according to the UD analysis, is on
the right conjunct, corresponding to the English \texttt{acl}, and the
first conjunct is attached to it using the \texttt{dep} label. Again,
the English-based model expects to find the root in the first of the
two clauses, and there is no overt complementiser to suggest
otherwise.

\subsection{Attention Patterns}

An analysis of the properties of the models underlying these findings
is beyond the scope of this paper, but preliminary checks of the
attention patterns show that successful models 
strongly attend to complementisers in the last two layers. As
SVO and VSO languages tend to have complementisers before subordinate 
clauses and SOV languages after
\citep{hawkins1990parsing}, fine-tuning biases models towards looking
for them in only one direction. The attention of subordinate-clause 
heads to main-clause heads is weaker, presumably due to higher
lexical variety in that position.




\section{Related Work}

Both aspects of our analysis -- subordinate-clause detection and the
study of word-order effects -- have been addressed but not in
conjunction and not in a multiple-source-language setting. Our study
extends previous approaches by providing a ZS \enquote{upper baseline}
derived from the study of the performance of several monolingual models 
and then conducting a novel many-sources-to-many-target analysis of zero-shot
performance.

\citet{lin-etal-2019-open} test BERT on the auxiliary-classification
task (main vs.\ subordinate clause) as part of their investigation of
BERT's linguistic knowledge. \citet{ronnqvist-etal-2019-multilingual}
extend this analysis to the multilingual setting with a focus on
Nordic languages.

Word-order differences have been shown to impact the performance of
English-based cross-lingual models, especially in the domain of
syntactic parsing \citep{ahmad-etal-2019-difficulties} and with tasks
that rely on syntactic information
\citep{liu2020importance,arviv-etal-2021-relation}, while reordering
has been long known to be an efficient preprocessing step in syntactic
transfer \citep{rasooli-collins-2019-low} and machine translation,
both statistical \citep{wang-etal-2007-chinese} and neural
\citep{chen-etal-2019-neural}.

\section{Conclusion}

We extend previous work on syntactic capabilities of BERT, mostly
focusing on English, by providing a more comprehensive analysis of its
performance on the task of subordinate-clause detection in multiple
languages and language pairs in the zero-shot setting. We show that
the performance of single-language models is uneven across languages:
East and Southeast Asian languages with less rigid boundaries between
POS categories and coordination and subordination prove harder to
model. We also show that mBERT's performance in the zero-shot setting,
while being largely correlated with the size of the pre-training and
fine-tuning corpora, with Russian being the best source language across
the board, is well aligned with the word-order typology: language
pairs with congruent word orders demonstrate better results, with both
SVO and SOV orders having higher in-group than across-group
accuracies. A~single pair of VSO languages in the data further
corroborates this finding, showing that the verb-final order is not
important \textit{per~se}.

The clause-initial position of complementisers in VSO languages partly
blurs this effect and helps SVO languages with large training corpora
serve as good sources for fine-tuning, but even Russian and English
fail on SOV languages, where complementisers tend to be postposed and
dependent-clause predicates never appear in the sentence-final
position. This shows that at least for some tasks, training on a
single source language is not enough. Moreover, our results from
single-language modelling seem to indicate that even superficially
simple syntactic tasks vary in difficulty across languages, which
imposes a hard limit on how well cross-lingual projection can perform.

\section*{Acknowledgements}

We thank Lilja Maria S\ae b{\o} and Chih-Yi Lin for their help with
the analysis of Mandarin Chinese data and Dojun Park for his help with the
analysis of Korean.


\clearpage

\bibliography{anthology,custom}
\bibliographystyle{acl_natbib}

\clearpage

\appendix

\section{Appendix}\label{sec:appendix}

\subsection{Abbreviations}\label{ssec:abbreviations}

\begin{itemize}
    \item[af] Afrikaans
    \item[ar] Standard Arabic
    \item[cs] Czech
    \item[cy] Welsh
    \item[de] German
    \item[en] English
    \item[es] Spanish
    \item[eu] Basque
    \item[ga] Irish
    \item[fi] Finnish
    \item[fr] French
    \item[he] Hebrew
    \item[hi] Hindi
    \item[hy] Eastern Armenian
    \item[id] Indonesian
    \item[is] Icelandic
    \item[it] Italian
    \item[ja] Japanese
    \item[ko] Korean
    \item[pl] Polish
    \item[pt] Portuguese
    \item[ru] Russian
    \item[sv] Swedish
    \item[th] Thai
    \item[tr] Turkish
    \item[vi] Vietnamese
    \item[zh] Mandarin Chinese
\end{itemize}

\subsection{Corpora}\label{ssec:corpora}

In addition to the Parallel Universal Dependencies collection \citep{zeman-etal-2017-conll},
the following corpora were used to train and/or validate models:

\begin{itemize}
    \item Afribooms: UD Afrikaans-AfriBooms,\\ \url{https://github.com/UniversalDependencies/UD_Afrikaans-AfriBooms}
    \item ArmTDP: Universal Dependencies treebank for Eastern Armenian,\\ \url{https://github.com/UniversalDependencies/UD_Armenian-ArmTDP}
    \item BDT: Basque UD treebank,\\ \url{https://github.com/UniversalDependencies/UD_Basque-BDT}
    \item CCG: Corpws Cystrawennol y Gymraeg (Syntactic Corpus of Welsh),\\ \url{https://github.com/UniversalDependencies/UD_Welsh-CCG}
    \item EWT: Universal Dependencies English Web Treebank,\\ \url{https://github.com/UniversalDependencies/UD_English-EWT}
    \item GSD (French): UD French GSD,\\ \url{https://github.com/UniversalDependencies/UD_French-GSD} \citep{guillaume2019conversion}
    \item GSD (Japanese): UD Japanese Treebank,\\ \url{https://github.com/UniversalDependencies/UD_Japanese-GSD}
    \item GSD (Korean): Google Korean Universal Dependency Treebank,\\ \url{https://github.com/UniversalDependencies/UD_Korean-GSD} \citep{chun-etal-2018-building}
    \item GSD (Mandarin): Traditional Chinese Universal Dependencies Treebank,\\ \url{https://github.com/UniversalDependencies/UD_Chinese-GSD}
    \item GSD (Spanish): Spanish UD treebank,\\ \url{https://github.com/UniversalDependencies/UD_Chinese-GSD}
    \item HDT: UD version of the Hamburg Dependency Treebank,\\ \url{https://github.com/UniversalDependencies/UD_German-HDT} \citep{borges-volker-etal-2019-hdt}
    \item HDTB: Hindi Universal Dependency Treebank,\\ \url{https://github.com/UniversalDependencies/UD_Hindi-HDTB} \citep{bhathindi}
    \item HTB: Universal Dependencies Corpus for Hebrew, \\\url{https://github.com/UniversalDependencies/UD_Hebrew-HTB} \citep{tsarfaty-2013-unified}
    \item IDT: Irish UD Treebank,\\ \url{https://github.com/UniversalDependencies/UD_Irish-IDT}
    \item KENET: Turkish-Kenet UD Treebank,\\ \url{https://github.com/UniversalDependencies/UD_Turkish-Kenet}
    \item PADT: UD version of the Prague Arabic Dependency Treebank,\\ \url{https://github.com/UniversalDependencies/UD_Arabic-PADT} \citep{padt}
    \item PDT: UD version of the Prague Dependency Treebank,\\ \url{https://github.com/UniversalDependencies/UD_Czech-PDT} \citep{pdt}
    \item Syntagrus: SynTagRus Dependency Treebank,\\ \url{https://github.com/UniversalDependencies/UD_Russian-SynTagRus}
    \item VTB: UD version of the VLSP constituency treebank,\\ \url{https://github.com/UniversalDependencies/UD_Vietnamese-VTB} \citep{nguyen-etal-2009-building}
\end{itemize}

\subsection{Single-language model results}\label{ssec:single-language-models}

The results attained by the models fine-tuned and tested on the same language are shown 
in Table~\ref{tab:single-lang-full}. See \S~\ref{ssec:corpora} for the details about the 
train and test corpora.

\begin{table*}[h!]
\centering
\resizebox{\textwidth}{!}{
\begin{tabular}{@{}llllllllllllll@{}}
\toprule
Language & Train corpus & Test corpus & Model & \#Train & \#Test & \begin{tabular}[c]{@{}l@{}}Main-\\ Main\end{tabular} & \begin{tabular}[c]{@{}l@{}}Main-\\ Sub\end{tabular} & \begin{tabular}[c]{@{}l@{}}Sub-\\ Main\end{tabular} & \begin{tabular}[c]{@{}l@{}}Sub-\\ Sub\end{tabular} & Acc. \\
\cmidrule(r){1-3}
\cmidrule(lr){4-4}
\cmidrule(lr){5-6}
\cmidrule(lr){7-10}
\cmidrule(l){11-11}
Mandarin & GSD-Train & PUD & mBERT & 3196 & 736 & 556 & 180 & 122 & 1364 & 86.4 \\
Mandarin & GSD-Train & PUD & \begin{tabular}[c]{@{}l@{}}HFL-\\BERT-\\ WWM\end{tabular} & 3196 & 736 & 570 & 166 & 85 & 1401 & 88.7 \\
Vietnamese & VTB-Train & VTB-Dev & mBERT & 1105 & 619 & 510 & 109 & 90 & 1283 & 90 \\
Korean & GSD-Train & PUD & mBERT & 2201 & 618 & 603 & 15 & 149 & 936 & 90.4 \\
Arabic & PADT-Train & PUD & mBERT & 3755 & 520 & 436 & 84 & 31 & 752 & 91.2 \\
Hindi & HDTB-Train & PUD & mBERT & 5167 & 565 & 506 & 59 & 32 & 831 & 93.6 \\
German & HDT-Train & PUD & mBERT & 55938 & 441 & 427 & 14 & 49 & 578 & 94.1 \\
Armenian & \begin{tabular}[c]{@{}l@{}}ArmTDP-\\ Train\end{tabular} & \begin{tabular}[c]{@{}l@{}}ArmTDP-\\ Dev\end{tabular} & mBERT & 1165 & 149 & 145 & 4 & 21 & 269 & 94.3 \\
Turkish & KENET-Train & PUD & mBERT & 6784 & 731 & 653 & 78 & 25 & 1338 & 95.1 \\
Welsh & CCG-Train & CCG-Dev & mBERT & 377 & 341 & 315 & 26 & 27 & 824 & 95.6 \\
Indonesian & GSD-Train & PUD & mBERT & 2770 & 572 & 553 & 19 & 42 & 923 & 96 \\
Basque & BDT-Train & BDT-Dev & mBERT & 3181 & 1029 & 979 & 50 & 39 & 1758 & 96.9 \\
Spanish & GSD-Train & PUD & mBERT & 7247 & 548 & 513 & 35 & 5 & 824 & 97.1 \\
Irish & IDT-Train & IDT-Dev & mBERT & 2323 & 236 & 226 & 10 & 8 & 441 & 97.4 \\
English & EWT-Train & PUD & \begin{tabular}[c]{@{}l@{}}BERT-\\ LARGE-\\ CASED\end{tabular} & 5968 & 556 & 529 & 27 & 4 & 915 & 97.9 \\
Hebrew & HTB-Train & HTB-Dev & mBERT & 2342 & 206 & 201 & 5 & 4 & 297 & 98.2 \\
Afrikaans & \begin{tabular}[c]{@{}l@{}}Afribooms-\\ Train\end{tabular} & \begin{tabular}[c]{@{}l@{}}Afribooms-\\ Train\end{tabular} & mBERT & 643 & 97 & 96 & 1 & 2 & 142 & 98.8 \\
French & GSD-Train & PUD & mBERT & 7712 & 572 & 563 & 9 & 6 & 956 & 99 \\
Japanese & GSD-Train & PUD & \begin{tabular}[c]{@{}l@{}}TOHOKU-\\ BERT-\\ LARGE\end{tabular} & 5101 & 844 & 838 & 8 & 18 & 2090 & 99.1 \\
Czech & PDT-Train & PUD & mBERT & 26277 & 504 & 502 & 2 & 3 & 779 & 99.6 \\
Russian & \begin{tabular}[c]{@{}l@{}}Syntagrus-\\ Train\end{tabular} & PUD & mBERT & 32851 & 595 & 593 & 2 & 2 & 961 & 99.7 \\ \bottomrule
\end{tabular}}
\caption{Performance of single-language models across languages. \#Train and \#Test denote the number of sentences in the train and test corpus respectively. In the \enquote{Main-Main}, \enquote{Main-Sub}, \enquote{Sub-Main}, and \enquote{Sub-Sub} columns, the part before the hyphen is the gold label of a predicate (main/subordinate clause) and the second part is the guessed label. Acc: Accuracy.}\label{tab:single-lang-full}
\end{table*}


\end{document}